\documentclass[11pt,a4paper]{article}
\usepackage[hyperref]{emnlp-ijcnlp-2019}
\usepackage{times}
\usepackage{latexsym}

\usepackage{url}
\usepackage{colortbl}
\usepackage{soul}
\usepackage{nicefrac}

\newcommand{\eg}{\textit{e}.\textit{g}., }

\usepackage[T1]{fontenc}
\usepackage{tgtermes}
\usepackage{amssymb}
\newcommand{\mathbold}[1]{\ensuremath{\boldsymbol{\mathbf{#1}}}}
\usepackage[ttdefault=true]{AnonymousPro}

\usepackage{amsfonts}
\usepackage{amsmath}
\usepackage{amsthm}
\usepackage{nicefrac}
\usepackage{bm}

\usepackage[english]{babel}
\usepackage[parfill]{parskip}
\usepackage{afterpage}
\usepackage{enumitem}
\usepackage{framed}
\usepackage{xspace}

\usepackage{lineno}

\newcounter{parcount}

\usepackage{graphicx}
\usepackage[labelfont=bf]{caption}
\usepackage{subcaption}

\usepackage{booktabs, array}

\usepackage{listings}
\usepackage{fancyvrb}
\fvset{fontsize=\small}

\usepackage[nameinlink]{cleveref}
\creflabelformat{equation}{#2\textup{#1}#3}  % <- remove parenthesis from equations

\newcommand{\g}{\,|\,}

\newcommand{\nestedmathbold}[1]{{\mathbold{#1}}}

\newcommand{\mbe}{\nestedmathbold{e}}

\newcommand{\mbh}{\nestedmathbold{h}}

\newcommand{\mbs}{\nestedmathbold{s}}
\newcommand{\mbt}{\nestedmathbold{t}}

\newcommand{\mbx}{\nestedmathbold{x}}
\newcommand{\mby}{\nestedmathbold{y}}

\newcommand{\mbH}{\nestedmathbold{H}}

\usepackage{multirow}
\newcommand{\expA}{\textit{expEn}}
\newcommand{\expB}{\textit{expLatin}}
\newcommand{\expC}{\textit{expMix}}
\newcommand{\expD}{\textit{expSOV}}

\newcommand{\AB}[1]{\textcolor{red}{[#1]}}

\newcommand{\COMMENT}[1]{}

\definecolor{fixred}{HTML}{ef3b2c}
\definecolor{LightGrey}{HTML}{d9d9d9}
\definecolor{svo}{HTML}{54278f}
\definecolor{sov}{HTML}{006d2c}
\aclfinalcopy % Uncomment this line for the final submission

\title{Zero-shot Dependency Parsing with \\  Pre-trained Multilingual Sentence Representations}

\author{Ke Tran\thanks{\;\;Work done prior to joining Amazon.} \\
  Amazon Alexa AI \\
  {\tt trnke@amazon.com} \\\And
  Arianna Bisazza\thanks{\;\;Work done while at Leiden University. Both authors contributed equally.} \\
  University of Groningen \\
  {\tt a.bisazza@rug.nl} \\}

\date{}

\begin{document}
\maketitle
\begin{abstract}
  We investigate whether off-the-shelf deep bidirectional sentence representations \cite{Devlin:2018} trained on a massively multilingual corpus (multilingual BERT) enable the development of an unsupervised universal dependency parser.
  This approach only leverages a mix of monolingual corpora in many languages and does not require any translation data making it applicable to low-resource languages.
  In our experiments we outperform the best CoNLL 2018 language-specific systems in all of the shared task's six truly low-resource languages while using a single system.
  However, we also find that (i) parsing accuracy still varies dramatically when changing the training languages and (ii) in some target languages zero-shot transfer fails under all tested conditions, raising concerns on the `universality' of the whole approach.
\end{abstract}

\section{Introduction}
Pretrained sentence representations \cite{Howard:2018,Radford:2018,peters:2018,Devlin:2018} have recently set the new state of the art in many language understanding tasks \cite{GLUE:2018}.
An appealing avenue for this line of work is to use a mix of training data in several languages and a shared subword vocabulary leading to general-purpose multilingual representations.
In turn, this opens the way to a number of promising cross-lingual transfer techniques that can address the lack of annotated data in the large majority of world languages.

In this paper, we investigate whether deep bidirectional sentence representations \cite{Devlin:2018} trained on a massively multilingual corpus (m-BERT) allow for the development of a universal dependency parser that is able to parse sentences in a diverse range of languages without receiving {\em any supervision} in those language. Our parser is fully lexicalized, in contrast to a successful approach based on delexicalized parsers \cite{Zeman:2008,mcdonald:2011}. %in the literature.
Building on the delexicalized approach, previous work employed additional features such as %WALS \cite{wals}
typological properties \cite{naseem:2012}, syntactic embeddings \cite{duong-EtAl:2015:CoNLL}, and cross-lingual word clusters \cite{Tackstrom:2012} to boost parsing performance.
More recent work by \citet{Ammar:2016,Guo:2016} requires translation data for projecting word embeddings into a shared multilingual space.

Among lexicalized systems in CoNLL18, the top system \cite{che-etal-2018-towards} utilizes contextualized vectors from ELMo. However, they train each ELMo for each language in the shared task. While their approach achieves the best LAS score on average, for low resource languages, the performance of their parser lags behind other systems that do not use pre-trained models \cite{conll2018}.
By contrast, we build our dependency parser on top of general-purpose context-dependent word representations pretrained on a multilingual corpus. This approach does not require any translation data making it applicable to truly low-resource languages (\S\ref{ssec:results}).
While m-BERT's training objective is inherently monolingual (predict a word in language $\ell$ given its sentence context, also in language $\ell$), we hypothesize that cross-lingual syntactic transfer occurs via the shared subword vocabulary and hidden layer parameters.
Indeed, on the challenging task of universal dependency parsing from raw text, we outperform by a large margin the best CoNLL18 language-specific systems \cite{conll2018} on the shared task's truly low-resource languages while using a single system.

The effectiveness of m-BERT for cross-lingual transfer of UD parsers has also been demonstrated in concurrent work by \newcite{Wu19} and \newcite{Kondratyuk19}.
While the former utilizes only English as the training language, the latter trains on a concatentation of all available UD treebanks.
We additionally experiment with three different sets of training languages beyond English-only and make interesting observations on the resulting large, and sometimes unexplicable, variation of performance among test languages.

\section{Model}
\label{sec:model}
We use the representations produced by BERT \cite{Devlin:2018} which is a self-attentive deep bidirectional network \cite{Vaswani:2017} trained with a masked language model objective.
Specifically we use BERT's multilingual cased version\footnote{\url{https://github.com/google-research/bert/blob/master/multilingual.md}} which was trained on the 100 languages with the largest available Wikipedias.
Exponentially smoothed weighting was applied to prevent high-resource languages from dominating the training data, and a shared vocabulary of 110k shared WordPieces \cite{Wu:2016} was used.
\COMMENT{
\begin{figure}[ht]
    \centering
    \includegraphics[scale=0.55]{figures/wo_pos.pdf}
    \caption{Parsing architecture. Grey boxes indicate BERT's parameters and output. White box indicates additional parameters of deep-biaffine layers. \AB{fig can now be removed}}
    \label{fig:net}
\end{figure}
}
For parsing we employ a modification of the graph-based dependency parser of \citet{Dozat16}.
We use deep biaffine attention to score arcs and their label from the head to its dependent.
While our label prediction model is similar to that of \citet{Dozat16}, our arc prediction model is a globally normalized model which computes partition functions of non-projective dependency structures using Kirchhoff's Matrix-Tree Theorem \cite{koo-EtAl:2007:EMNLP}.

Let $\mbx = w_1, w_2, \dots, w_n$ be an input sentence of $n$ tokens, which are given by the gold segmentation in training or by an automatic tokenizer in testing (\S\ref{ssec:data}).
To obtain the m-BERT representation of $\mbx$,
we first obtain a sequence $\mbt=t_1, \dots, t_m$ of $m\geq n$ subwords from $\mbx$ using the WordPiece algorithm.
Then we feed $\mbt$ to m-BERT and extract the representations  $\mbe_1, \dots, \mbe_m$ from the last layer.

If word $w_i$ is tokenized into $(t_j, \dots, t_k)$ then the representation $\mbh_i$ of $w_i$ is computed as the mean of $(\mbe_j,\dots,\mbe_k)$.

The arc score is computed similar to \citet{Dozat16}:
\begin{align}
\mbs^{\text{(arc)}} &= \mbox{\small\texttt{DeepBiaffine}}(\mbH^{\textrm{(arc-head)}}, \mbH^{\textrm{(arc-dep})})
\end{align}
The $\log$ probability of the dependency tree $\mby$ of $\mbx$ is given by
\begin{align}
\log p(\mby\g\mbx) &= \sum_{(h,c) \in \mby}\mbs^{\text{(arc)}}[h,c] - \log Z(\mbx)\label{eq:mt}
\end{align}
where $Z(\mbx)$ is the partition function. Our objective function for predicting dependency arcs therefore is globally normalized.
We compute $Z(\mbx)$ via matrix determinant \cite{koo-EtAl:2007:EMNLP}.
In our experiments, we find that training with a global objective is more stable if the score $\mbs^{\text{(arc)}}[h,c]$ is locally normalized\footnote{We use {\small\texttt{log\_softmax}}($\mbs^{\text{(arc)}}$) in place of $\mbs^{\text{(arc)}}$ in equation~\ref{eq:mt}.} such that $\sum_h \exp(\mbs^{\text{(arc)}}[h,c]) = 1$.
During training, we update both m-BERT and parsing layer parameters.

\section{Experiments}
\label{sec:experiments}
While most previous work on parser transfer, including the closely related \cite{duong-EtAl:2015:CoNLL} relies on gold tokenization and POS tags, we adopt the more realistic scenario of parsing from \textit{raw text} \cite{conll2018} and adopt the automatic sentence segmenter and tokenizer provided as baselines by the shared task organizers.

\subsection{Data}
\label{ssec:data}

We use the UDpipe-tokenized test data\footnote{Preprocessed data available at \url{http://hdl.handle.net/11234/1-2899}}  \cite{straka-strakova:2017:K17-3}
and the CoNLL18 official %\texttt{conll18\_ud\_eval.py}
script %\footnote{\url{http://universaldependencies.org/conll18/conll18_ud_eval.py}}
for evaluation. Gold tokenization is only used for the training data, while POS information is never used. All of our experiments are carried out on the Universal Dependencies (UD) corpus version 2.2 \cite{ud2.2} for a fair comparison with previous work.

While our sentence representations are always initialized from m-BERT, we experiment with four sets of parser training (i.e. fine-tuning) languages, namely: \expA\ only English (200K words); \expB\ a mix of four Latin-script European languages: English, Italian, Norwegian, Czech (50K each, 200K in total); \expD\ a mix of two SOV languages: Hindi and Korean  (100K each, 200K in total); \expC\ a larger mix of eight languages including different language families and scripts: English, Italian, Norwegian, Czech, Russian, Hindi, Korean, Arabic (50K each, 400K in total).
For high resource languages that have more than one treebank, we choose the treebank that has the best LAS score in ConLL18 for training and the lowest LAS score for zero-shot evaluation.

\subsection{Training details}
\label{ssec:training}
Similar to \citet{Dozat:2017}, we use a neural network  output size of 400 for arc prediction and 100 for label prediction. We use the Adam optimizer with learning rate  $5e^{-6}$ to update the parameters of our models. The model is evaluated every 500 updates and we stop training if the score LAS does not  increase in ten consecutive validations.

\begin{table*}[ht]
\centering\small
\begin{tabular}{@{}l l c c c c c c c r@{}}
\toprule
 &  & \multicolumn{4}{c}{m-BERT based} && \multicolumn{2}{c}{State of the art} & \\
target & tbk-code & \expA & \expB & \expD & \expC && Stanford & CoNLL18 & \#TrWrds\\
\midrule
Russian & \textcolor{svo}{\texttt{ru\_syntagrus}} &  59.53 & 73.13 & 34.44 & \cellcolor{LightGrey}{81.91} && 91.20 & 92.48 & 872 K \\
Hindi & \textcolor{sov}{\texttt{hi\_hdtb}} & 32.94 & 33.75 & \cellcolor{LightGrey}{88.51} & \cellcolor{LightGrey}{85.66} && 91.65 & 92.41 & 281 K \\
Italian & \textcolor{svo}{\texttt{it\_isdt}} & 75.45 & \cellcolor{LightGrey}{89.59} & 25.95 &  \cellcolor{LightGrey}{89.44} && 90.51 & 92.00 & 276 K\\
Norwegian & \textcolor{svo}{\texttt{no\_nynorsk}} & 72.09 & \cellcolor{LightGrey}{86.01} & 33.93 & \cellcolor{LightGrey}{85.11} && 89.58 & 90.99 & 245 K \\
Czech & \textcolor{svo}{\texttt{cs\_pdt}} & 59.97 & \cellcolor{LightGrey}{84.91} & 34.31 & \cellcolor{LightGrey}{84.36} && 89.63 & 89.63 & 1,173 K \\
Finnish & \textcolor{svo}{\texttt{fi\_tdt}} & 50.65 & 61.13 & 40.12 & 62.29 && 86.33 & 88.73 & 163 K\\
Persian & \textcolor{sov}{\texttt{fa\_seraji}} & 44.34  & 56.39 & 24.77 & 56.92 && 86.55 & 88.11 & 121 K \\
Korean & \textcolor{sov}{\texttt{ko\_kaist}} & 33.67 & 38.87 & \cellcolor{LightGrey}{84.39} & \cellcolor{LightGrey}{81.73} && 86.58 & 86.91 & 296 K \\
English & \textcolor{svo}{\texttt{en\_ewt}} & \cellcolor{LightGrey}{84.64} & \cellcolor{LightGrey}{82.38} & 30.03 & \cellcolor{LightGrey}{81.65} && 83.80 & 84.57 & 205 K \\
Urdu & \textcolor{sov}{\texttt{ur\_udtb}} & 23.46 & 23.94 & 65.21 & 63.06 && 82.58 & 83.39 & 109 K\\
Japanese & \textcolor{sov}{\texttt{ja\_gsd}} & 12.92 & 12.65 & 19.25 & 24.10 && 78.48 & 83.11 & 162 K\\
Hungarian & \texttt{hu\_szeged} & 52.72 & 61.11 & 39.65 & 61.11 && 78.58 & 82.66 & 20 K \\
German & \texttt{de\_gsd} & 68.30 & 70.93 & 36.30 & 70.93 && 79.17 & 80.36 & 264 K \\
Swedish & \textcolor{svo}{\texttt{sv\_pud}} & 76.02 & 78.71 & 37.58 & 78.70 && 78.39 & 80.35 & -- K \\
Arabic & \texttt{ar\_padt} & 34.55 & 50.20 & 12.26 & \cellcolor{LightGrey}{68.20} && 76.99 & 77.06 & 224 K \\
French & \textcolor{svo}{\texttt{fr\_spoken}} & 54.12 & 59.70 & 16.06 & 59.54 && 69.56 & 75.78 & 15 K \\
Vietnamese & \textcolor{svo}{\texttt{vi\_vtb}} & 29.72 & 30.09 & 14.13 & 29.71 && 47.56 & 55.22 & 20 K\\
Tamil & \textcolor{sov}{\texttt{ta\_ttb}} & 18.09 & 25.79 & 29.64 & 32.78 && -- & -- & 5 K\\
Telugu & \textcolor{sov}{\texttt{te\_mtg}} & 54.47 & 63.06 & 61.68 & 64.03 && -- & -- & 5 K\\
\midrule
Faroese* & \textcolor{svo}{\texttt{fo\_oft}} & 58.28 & 61.71 & 36.27 & \bf 61.98 && 41.54 & 49.43 & 0 K \\
Upper Sorbian* & \textcolor{sov}{\texttt{hsb\_ufal}} &  36.66 & \bf 49.90 & 23.90 & 49.74 && 23.61 & 46.42 & 0 K \\
Breton & \texttt{br\_keb} & 45.16 & 51.85 & 22.49 & \bf 52.62 && 11.25 & 38.64 & 0 K \\
Armenian & \textcolor{sov}{\texttt{hy\_armtdp}} & 40.20 & 55.44 & 41.91 & \bf 58.95 && 31.47 & 37.01 & 1 K \\
Kazakh & \textcolor{sov}{\texttt{kk\_ktb}} & 33.56 & 40.18 & 40.18 & \bf 44.56 && 26.25 & 31.93 & 1 K \\
Buryat* & \textcolor{sov}{\texttt{bxr\_bdt}} & 19.19 & 20.90 & 22.94 & \bf 23.11 && 12.47 & 19.53 & 0 K \\
\midrule
avg(lowRes) &  & 39.41 & 47.26 &  31.28 & 48.45 && 24.43 & 37.16 & \\
\bottomrule
\end{tabular}
\caption{LAS scores of our parser in the raw text setup. Languages not in m-BERT's training corpus are marked with *. \textcolor{svo}{SVO} and \textcolor{sov}{SOV} languages are indicated by \textcolor{svo}{purple} and \textcolor{sov}{green} respectively. Stanford and CoNLL18's best submitted systems are provided as representative state-of-the-art supervised systems. \#TrWrds = Total training data made available at CoNLL18. %, rounded to the nearest thousand words.
The amount of training used in each experiment is specified in \S\ref{ssec:data}. Training languages for each experiment are highlighted in grey.}
\label{tb:xscores}
\end{table*}%

\subsection{Results}
\label{ssec:results}
To put our results into perspective, we report the accuracy of the best CoNLL18 system for each language and that of the Stanford system submitted at the same evaluation \cite{qi2018universal}. The latter is also based on the deep biaffine parser of \citet{Dozat16}, it does not use ensembles and was ranked $2^{\text{nd}}$ on official evaluation metric LAS\footnote{Updated results at \url{https://stanfordnlp.github.io/stanfordnlp/performance.html} March, 2018}.
Both these parsers receive supervision in most of the languages, therefore comparison to our parser is only fair for the low-resource languages where training data is not available (or negligible, i.e. less than 1K tokens).

Results for a subset of UD languages are presented in Table~\ref{tb:xscores}. Beside  common European languages, we choose languages with different writing scripts than those presented in the parser training data. We also include SOV (\eg Korean, Persian) and VSO (\eg Arabic, Breton) languages.
Parser training languages for each experiment are highlighted in grey in Table~\ref{tb:xscores}.

In the high resource setting, there is a considerable gap between zero-shot and supervised parsers with Swedish as an exception (slightly better than Stanford's parser and 2 points below  CoNLL18).
By contrast, the benefit of multilingual transfer becomes evident in the low resource setting.
Here, most CoNLL18 systems including Stanford's use knowledge of each target language to customize the parser, \eg to choose the optimal training language(s).
Nevertheless, our single parser trained on the largest mix of languages (\expC) beats the best CoNLL18 language-specific systems on all six languages, even though three of these languages are not represented in m-BERT's training data\footnote{This is possible because their sub-words are in BERT's vocabulary due other similar languages in training data.}. This result highlights the advantage of multilingual pre-trained model in the truly low resource scenario.

We notice the poor performance of our parser on  spoken French  in comparison to other European languages.
While there is sufficient amount of Wikipedia text for French, it seems that zero-shot parsing on a different domain remains a challenge even with a large pre-trained model.

\section{Analysis}
\label{sec:analysis}
By varying the set of parser training languages we analyze our results with respect to two factors: parser training language diversity and word order similarity.

\subsection{Training language diversity}
\label{ssec:diversity}
Increasing language diversity (\expA$\rightarrow$\expB\ and \expB$\rightarrow$\expC) leads to improvements in most test languages, even when the total amount of training data is fixed (\expA$\rightarrow$\expB). The only exceptions are the languages for which training data is reduced (English in \expB) or becomes a smaller proportion of the total training data (Czech, Italian, Norwegian in \expC), which confirms previous findings \cite{Ammar:2016}. Swedish and Upper Sorbian being related to Norwegian and Czech respectively also lose some accuracy in \expC. On the other hand, newly included languages  (Czech, Italian, Norwegian in \expB\ and Arabic, Hindi, Korean, Russian in \expC) show the biggest improvements, which was also expected.

More interestingly, some large gains are reported for languages that are unrelated from all training languages of \expB. %and even have non-Latin scripts.
We hypothesize that such languages (Arabic, Armenian, Hungarian) may benefit from an exposure of the parser to a more diverse set of word orders (\S\ref{ssec:word_order}). For instance, Arabic being head initial is closer to Italian than to English in terms of word order.

Actual language relatedness does not always play a clear role: For instance, Upper Sorbian seems to benefit largely from its closeness to Czech in \expB\ and \expC, while Faroese (related to Norwegian) does not improve as much.

In summary, language diversity in training is clearly a great asset. However, there is a large variation in gains among test languages, for which language family relatedness can only offer a partial explanation.

\subsection{Training language typology}
\label{ssec:word_order}
Training on languages with similar typological features has been shown beneficial for parsing target languages in the delexicalized setup. In particular, word order similarities have been proved beneficial to select source languages for parsing model transfer \cite{naseem:2012,duong-EtAl:2015:CoNLL}.
Indeed, when Hindi and Korean are presented in \expD, we report %see a large boost in LAS scores
better LAS scores in various SOV languages (Japanese, Tamil, Urdu, Buryat) however some other SOV languages (Persian and Armenian) perform much worse than in \expB\  showing that word order is not a reliable criterion for training language selection.

Given these observations, we construct our largest training data (\expC) by merging all the languages of \expA, \expB, and \expD\ and adding two more languages with diverse word order profiles for which large treebanks exist, namely Russian and Arabic.

Concurrently to this work, \newcite{Lin19} have proposed an automatic method to choose the optimal transfer languages in various tasks including parsing, based on a variety of typological but also data-dependent features. We leave adoption of their method to future work.\footnote{Unfortunately at the time of writing we have not yet managed to use their released implementation.}

\subsection{Towards explaining transfer performance}
\label{ssec:subword}

Even when keeping the training languages fixed, for instance in \expC, we observe a large variation of zero-shot parsing transfer accuracy among test languages which does not often correlate with supervised parsing accuracy.
As an attempt to explain this variation we look at the overlap of test vocabulary with
(i) parser's training data vocabulary $\tau$ and
(ii) m-BERT's training data vocabulary.
Because m-BERT uses a subword vocabulary that also includes characters we resort to measuring the unsegmented word score $\eta$:
\begin{align}
    \tau &= 100 \times\nicefrac{|\texttt{type\_w}(D_\text{test}) \,\cap\, \texttt{type\_w}(D_\text{train})|}{|\texttt{type\_w}(D_\text{test})|}\nonumber\\
    \eta &= 100 \times\nicefrac{|\texttt{token\_g}(D_\text{test})|}{|\texttt{token\_w}(D_\text{test})|}\nonumber
\end{align}
where $\texttt{type\_w}(D)$ and $\texttt{token\_w}(D)$ are sets of WordPieces types and tokens in dataset $D$ respectively, and $\texttt{token\_g}(D)$ is the set of gold tokens in $D$ before applying WordPieces.
A higher $\eta$ indicates a less segmented text.

To account for typological features, we also plot the average syntactic similarity $\bar{\sigma}$ of each test language to the eight \expD\ training languages as computed by the URIEL database\footnote{Specifically, we compute $1-d$ where $d$ is the pre-computed {\em syntactic} distance in \texttt{lang2vec}.}
\cite{uriel}.

\begin{figure}[ht]
\centering
\includegraphics[scale=.70]{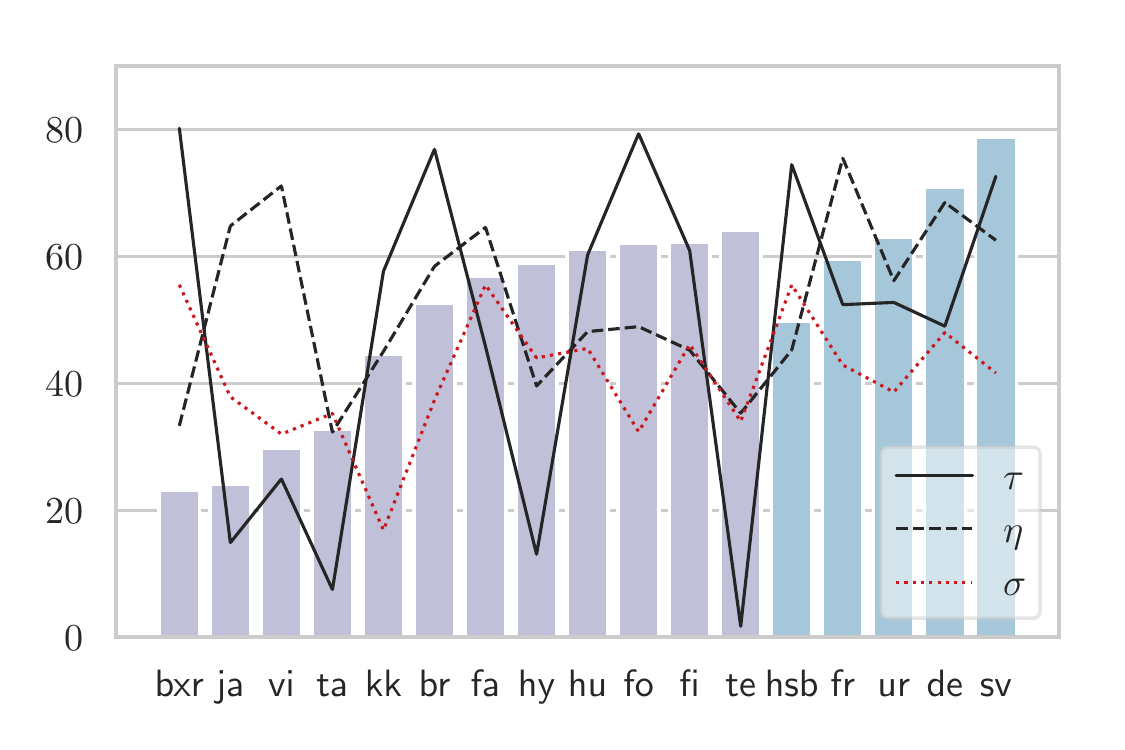}
\caption{Relationship between parsing accuracy (\expC), parser training-test vocabulary overlap $\tau$, \mbox{m-BERT} unsegmented word score $\eta$, and average typological syntactic similarity $\bar{\sigma}$. Purple bar indicates there is no language that belongs to the same family presented in training data. Languages in the training set of \expC \ are not shown.}
\label{fig:vocab_stats}
\end{figure}
We observe a correlation between LAS, $\eta$ and $\tau$ for test languages that have a relative in the training data, like Urdu and Hindi. For test languages that belong to a different family than all training languages, no correlation appears. A similar observation is also reported by \newcite{pires-etal-2019-multilingual}: namely, they find that the performance of cross-lingual named entity recognition with \mbox{m-BERT} is largely independent of vocabulary overlap.

Although typological features have been shown to be useful when incorporated into the parser \citep{naseem:2012,Ammar:2016}, we do not find a clear correlation between $\bar{\sigma}$ and LAS in our setup.
Thus none of our investigated factors can explain transfer performance in a systematic way.

\subsection{Language outliers}

While massively pre-trained language models promise a more inclusive future for NLP, we find it important to note that cross-lingual transfer performs very badly for some languages.

For instance, in our experiments, Japanese and Vietnamese stand out as strikingly negative outliers.
\newcite{Wu19} also report a very low performance on Japanese in their zero-shot dependency parsing experiments.\footnote{They do not report parsing results for Vietnamese.} In \cite{Lin19} Japanese is completely excluded from the parsing experiments because of unstable results.

Japanese and Vietnamese are \textit{language isolates} in an NLP sense, meaning that they do not enjoy the presence of a closely related language among the high-resourced training languages.\footnote{The original definition of language isolate in linguistics is actually stronger: ``a language that has no known relatives, that is, that has no demonstrable phylogenetic relationship with any other language'' \cite{langisolates}
}
For this class of languages, transfer performance is overall very inconsistent and hard to explain.

\begin{table}[ht]
\centering
\begin{tabular}{l c c}
\toprule
& UDpipe & Gold \\
\midrule
\texttt{ko}$\rightarrow$\texttt{ja} & 14.96 & 20.04 \\
\texttt{ja}$\rightarrow$\texttt{ko} & 37.44 & 37.45 \\
\midrule
\end{tabular}
\caption{LAS scores when transferring between Korean and Japanese in two tokenization conditions.}
\label{tab:japkor}
\end{table}

The case of Japanese is particularly interesting for its relation to Korean. Family relatedness between these two languages is very controversial but their syntactic features are extremely similar.
To put our parser in optimal transfer conditions, we perform one last experiment by training only on Korean (all available data) and testing on Japanese, and vice versa. As shown in Table~\ref{tab:japkor},
Japanese performance becomes even lower in this setup.
We can also see that transferring in the opposite direction leads to a much better result, despite the fact that state-of-the-art supervised systems in these two languages achieve similar results (Japanese: 83.11, Korean: 86.92 by the best CoNLL18 systems).
To rule out the impact of unsupervised sentence and token segmentation, which may be performing particularly poorly on some languages, we retrain the parser with gold segmentation and find that it explains only a small part of the gap.

While \citet{pires-etal-2019-multilingual} hypothesize word order is the main culprit for the poor zero-shot performance for Japanese when transferring a POS-tagger from English, our experiments with Korean and Japanese show a different picture.

\section{Conclusions}
\label{sec:conclusions}
We have built a Universal Dependency parser on top of deep bidirectional sentence representations pre-trained on a massively multilingual corpus (m-BERT) without any need for parallel data, treebanks or other linguistic resources in the test languages.

Evaluated in the challenging scenario of parsing from raw text, our best parser trained on a mix of languages representing both language family and word order diversity
outperforms the best CoNLL18 language-specific systems on the six truly low-resource languages presented at the shared task.

Our experiments show that language diversity in the training treebank is a great asset for transfer to low-resource languages.
Moreover, the massively multilingual nature of m-BERT does not neutralize the impact of transfer languages on parsing accuracy, which is only partially explained by language relatedness and word order similarity.

Finally we have raised the issue of language outliers that perform very poorly in all our tested conditions and that, given our analysis, are unlikely to benefit even from automatic methods of transfer language selection \cite{Lin19}.

\section*{Acknowledgements}
Arianna Bisazza was funded by the the Netherlands Organization for Scientific Research (NWO) under project number 639.021.646.

\bibliography{emnlp-ijcnlp-2019}
\bibliographystyle{acl_natbib}
\end{document}